\documentclass[]{oppo}

\usepackage[toc,page,header]{appendix}

\usepackage{minitoc}

\usepackage[utf8]{inputenc} 
\usepackage[T1]{fontenc}    
\usepackage{hyperref}       
\usepackage{url}            
\usepackage{booktabs}       
\usepackage{amsfonts}       
\usepackage{nicefrac}       
\usepackage{microtype}      
\usepackage{xcolor}         
\usepackage{amsmath}
\usepackage{xspace}
\usepackage{graphicx}
\usepackage{multicol}
\newcommand{\method}{\textsc{Efficient Agents \xspace}}

\usepackage[most]{tcolorbox} 
\tcbset{
  aibox/.style={
    width=\linewidth,
    top=8pt,
    bottom=4pt,
    colback=black!5!white,
    colframe=ogreen,
    colbacktitle=ogreen,
    enhanced,
    center,
    attach boxed title to top left={yshift=-0.1in,xshift=0.15in},
    boxed title style={boxrule=0pt,colframe=white,},
  }
}
\newtcolorbox{AIbox}[2][]{aibox,title=#2,#1}

\newtcolorbox{promptbox}[2][Prompt]{
colback=black!5!white,
arc=5pt, 
boxrule=0.5pt,
fonttitle=\bfseries,
title=#1, 
before upper={\small}, fontupper=\fontfamily{ptm}\selectfont,
colframe=#2, 
}
\definecolor{ogreen}{RGB}{34, 139, 34}

\title{Efficient Agents: \\ Building Effective Agents While Reducing Cost}

\affiliation{OPPO AI Agent Team}

\abstract{
The remarkable capabilities of Large Language Model (LLM)-driven agents have enabled sophisticated systems to tackle complex, multi-step tasks, but their escalating costs threaten scalability and accessibility. This work presents the first systematic study of the efficiency-effectiveness trade-off in modern agent systems, addressing the critical need for cost-effective designs without sacrificing performance. We investigate three key questions: (1) How much complexity do agentic tasks inherently require? (2) When do additional modules yield diminishing returns? (3) How much efficiency can be gained through the design of efficient agent frameworks? Through an empirical analysis on the GAIA benchmark, we evaluate the impact of LLM backbone selection, agent framework designs, and test-time scaling strategies. Using the cost-of-pass metric, we quantify the efficiency-performance trade-off across these dimensions. Our findings inform the development of \method, a novel agent framework that has an optimal complexity to task requirements. \method retains 96.7\% of the performance of OWL, one leading open-source agent framework, while reducing operational costs from \$0.398 to \$0.228, resulting in a 28.4\% improvement in cost-of-pass. Our work provides actionable insights for designing efficient, high-performing agent systems, advancing the accessibility and sustainability of AI-driven solutions.
}

\date{\today}
\correspondence{Wangchunshu Zhou at \email{zhouwangchunshu@oppo.com}; He Zhu at \email{zhuhe@oppo.com}}
\checkdata[Code]{\url{https://github.com/OPPO-PersonalAI/OAgents}}

\begin{document}
\maketitle


\section{Introduction}
The ever-increasing reasoning and creation capabilities of Large Language Models (LLMs) have opened up a broad prospect for real-world applications. Researchers have developed numerous LLM-driven agent systems \citep{wu2023autogen,zhou2023agents,zhou2024agents2,zhou2023recurrentgpt,autogpt2025,zhu2025oagentsempiricalstudybuilding,zhu2025scalingtesttimecomputellm,hu2024infiagentdabenchevaluatingagentsdata,hu2024agents,shi2025taskcraftautomatedgenerationagentic} and created a large number of fascinating products capable of handling complex, multi-step tasks. However, this progress mirrors a familiar trajectory in NLP research: from BERT \citep{devlin2019bertpretrainingdeepbidirectional} to ChatGPT \citep{achiam2023gpt}, researchers consistently prioritize scaling up models to achieve breakthrough capabilities \citep{kaplan2020scalinglawsneurallanguage,hoffmann2022trainingcomputeoptimallargelanguage}, only later turning to optimize efficiency, cost, and environmental impact \citep{sanh2020distilbertdistilledversionbert,abdin2024phi4technicalreport}. This pattern has given rise to the critical subfield of efficient NLP, where researchers balance performance with practical constraints like inference latency, energy consumption, and economic viability. 
 
We argue that agent research has now reached a similar inflection point. While increasingly sophisticated agent architectures can solve remarkably complex problems, their costs scale prohibitively. Industry deployments reveal this tension starkly: cutting-edge agent products (e.g., DeepResearch \citep{openai_deep_research}, Manus \citep{manus}) demonstrate impressive capabilities but suffer from exorbitant operating costs due to explosive LLM call overhead. Some systems require hundreds of API calls per task, rendering them economically unsustainable despite their technical brilliance. This creates a fundamental bottleneck for real-world adoption, limiting both the scalability of applications and the accessibility of AI advancements.

Our work presents the first systematic study of the efficiency-effectiveness trade-off in modern agent systems. Through rigorous empirical analysis, we investigate 3 research questions: (1) How much complexity do agentic tasks truly require? (2) When do additional modules yield diminishing returns? (3) How much efficiency can be gained through the design of task-adaptive agent frameworks? By dissecting these relationships across the framework, we provide actionable insights for both researchers and practitioners. 

\begin{figure*}
    \centering
    \includegraphics[width=1.0\linewidth]{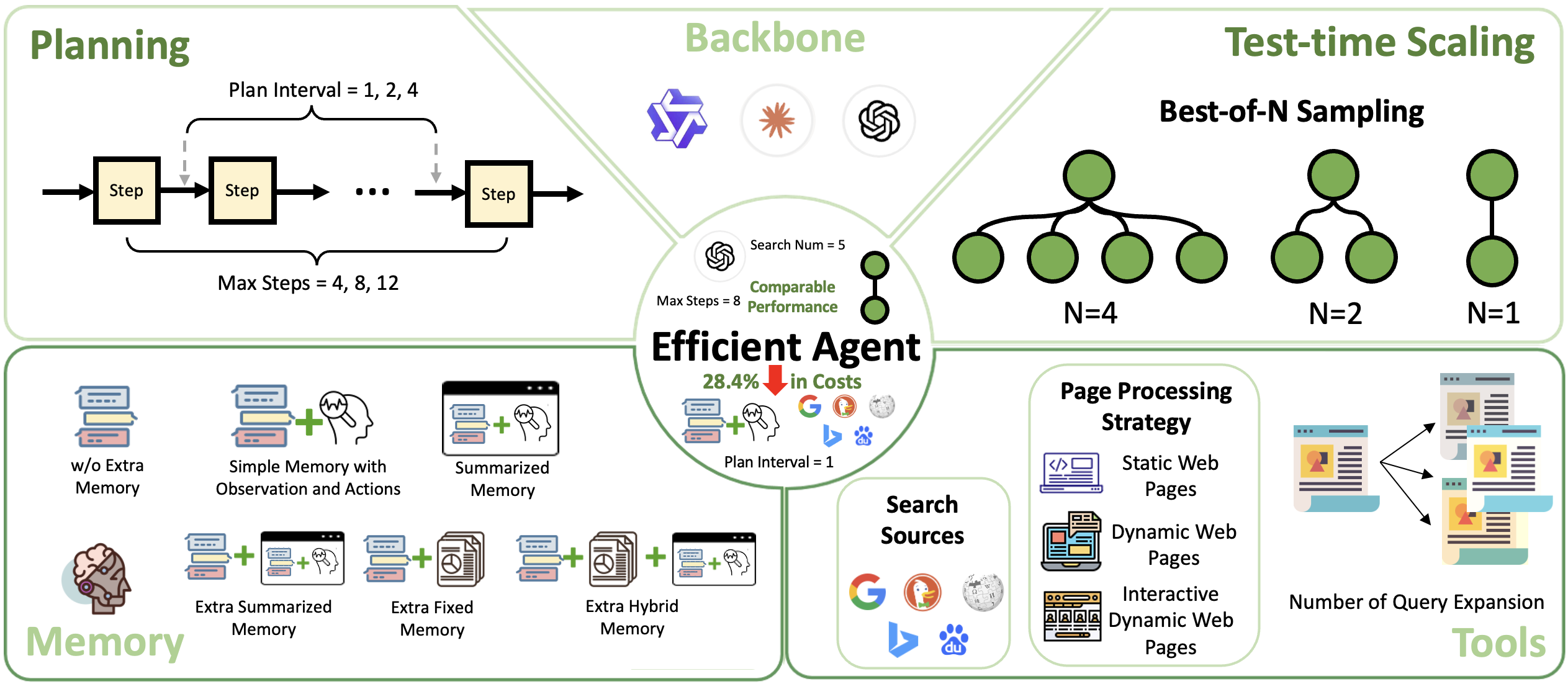}
    \caption{Evaluation of effectiveness and efficiency in agent system components. We adopt cost-of-pass as the metric to evaluate. We develop \method \ that optimizes cost while maintaining accuracy.}
    \label{fig:main}
\end{figure*}

We conduct an empirical study on the GAIA benchmark \citep{mialon2023gaiabenchmarkgeneralai} focusing on the efficiency-performance trade-off of agent systems by evaluating the impact of: (1) the choice of LLM backbones; (2) the designs of agent frameworks including planning, tool using, and memory; (3) test-time scaling and ensemble strategies. We adopt the cost-of-pass~\citep{erol2025costofpasseconomicframeworkevaluating} metric to compare and analyze the impact of different design choices on the efficiency-performance trade-off of LLM-based agents, as illustrated in Figure \ref{fig:main}. Based on the insight obtained in the aforementioned empirical study, we introduce \method, an agent framework optimized for achieving the best efficiency-performance trade-off that achieves a new state-of-the-art on the cost-of-pass metric on the GAIA benchmark. Specifically, 96.7\% of the performance of OWL~\citep{owl2025}, an open-source agent framework that achieves great performance on the GAIA benchmark, while reducing the cost from \$0.398 to \$0.228, leading to a relative improvement of 28.4\% in terms of cost-of-pass.

Our contribution could be summarized as follows:

\begin{itemize}
    \item We thoroughly analyze and summarize the factors that cause significant economic overhead in a generic LLM-based agent system.
    \item We propose \method, an efficient agent system where each component is selected based on prior analytical results, optimizing for efficiency while maintaining high effectiveness. Experimental results demonstrate that \method achieves 96.7\% of the performance of OWL while reducing costs by 28.4\%.
\end{itemize}

\section{Preliminaries}
\subsection{Setup}
Many factors can influence the effectiveness and efficiency of an agent system \citep{wang2024reasoning}. In this paper, we aim to conduct a comprehensive analysis from the perspective of agent systems. These factors encompass not only the backbone LLM itself but also the agent framework built around it, including planning mechanisms \citep{huang2024understandingplanningllmagents}, tool usage \citep{qin2024toollearningfoundationmodels} , and memory module \citep{zhang2024surveymemorymechanismlarge}. In addition, test-time scaling strategies \citep{snell2024scalingllmtesttimecompute} are also considered.
To this end, we evaluate these components on GAIA \citep{mialon2023gaiabenchmarkgeneralai}, a popular and challenging agent benchmark. This benchmark typically requires agents to perform complex reasoning to solve problems. By leveraging the benchmark, we can effectively assess the impact of individual components on overall performance and efficiency. Additionally, we compare several distinct agent frameworks to provide a broader perspective on their relative strengths. To accurately identify the impact of each component on efficiency and effectiveness, we established a default setup (listed in the Appendix) and varied one component at a time, testing to observe the effects.

\subsection{Metrics}
An ideal agent should achieve both high performance and computational efficiency. Therefore, in addition to accuracy, measured by pass@1 (solving the problem in one attempt) to evaluate effectiveness, we assess efficiency using the number of tokens taken by LLMs and associated costs. Notably, the cost of input tokens for APIs of LLMs is generally significantly lower than that of output tokens across many proprietary LLM providers. Accordingly, we compute these costs separately. The per-token pricing is sourced from official provider documentation as of May 2025. Furthermore, as different models or strategies may simultaneously impact performance and efficiency, we follow \citep{erol2025cost} and adopt the cost-of-pass metric to quantify model efficiency. 
The cost-of-pass metric, denoted as $v(m,p)$, represents the expected monetary cost of using a model $m$ to generate a correct solution for a problem $p$. It is computed as the ratio of the cost of a single inference attempt, $C_m(p)$, to the success rate, $R_m(p)$:
\[
v(m,p) = \frac{C_m(p)}{R_m(p)}
\]
Here, the cost of a single inference attempt, $C_m(p)$, is defined as:
\[
C_m(p) = n_{\text{in}}(m,p) \cdot c_{\text{in}}(m) + n_{\text{out}}(m,p) \cdot c_{\text{out}}(m)
\]
where $n_{\text{in}}(m,p)$ and $n_{\text{out}}(m,p)$ are the number of input and output tokens for model $m$ on problem $p$, respectively, and $c_{\text{in}}(m)$ and $c_{\text{out}}(m)$ are the per-token costs for input and output. The success rate $R_m(p)$ is estimated by the proportion of correct responses.
This metric represents the expected monetary cost of using a model to generate a correct solution for a problem, providing a comprehensive measure of a model's economic efficiency.

\section{On the Efficiency-Performance Trade-off of Agent Systems}

\subsection{Backbones}
\label{sec:large_reasoning_models}
Current Large Language Models acquire System-2 reasoning capabilities through reinforcement learning \citep{jaech2024openai,guo2025deepseek}, leveraging extended chain-of-thought \citep{wei2022chain} processes that often span thousands of tokens or more. While this approach significantly enhances reasoning performance, it also substantially increases computational costs and even leads to the phenomenon of overthinking \citep{chen2025think23overthinkingo1like}, which means excessive computational resources are allocated to simple problems during inference. To investigate this trade-off, we evaluate the performance and efficiency of several models with the same agent frameworks, including proprietary models such as GPT-4.1 \citep{gpt41}, o1 \citep{jaech2024openai} and Claude-3.7 \citep{Anthropic2025Claude37Sonnet}, open-source sparse models with MoE architecture such as Qwen3-235B-A22B \citep{yang2025qwen3} and Qwen3-30B-A3B \citep{yang2025qwen3}, also dense model such as QwQ-32B \citep{qwq32b}. The results are presented in Table \ref{tab:model}.
\begin{table*}[h]
\centering
\caption{Performance of various backbone LLMs on the GAIA dataset. We report metrics across the entire GAIA development set and its three difficulty levels (Level 1 to Level 3). Cost-of-pass is defined as infinity when accuracy is zero, signifying that no benefits can be obtained irrespective of the cost incurred.}
\resizebox{\textwidth}{!}{
\begin{tabular}{c|cccc|cccc|cccc|cccc}
\toprule
\multirow{3}{*}{Method} & \multicolumn{4}{c|}{Efficiency} & \multicolumn{4}{c|}{Effectiveness} & \multicolumn{8}{c}{Cost} \\
\cmidrule(lr){2-5} \cmidrule(lr){6-9} \cmidrule(lr){10-17}
& \multicolumn{4}{c|}{cost-of-pass$\downarrow$} & \multicolumn{4}{c|}{Acc./\%$\uparrow$} & \multicolumn{4}{c|}{Cost/\textdollar$\downarrow$} & \multicolumn{4}{c}{\#Tokens$\downarrow$} \\
\cmidrule(lr){2-5} \cmidrule(lr){6-9} \cmidrule(lr){10-13} \cmidrule(lr){14-17}
& all & l1 & l2 & l3 & all & l1 & l2 & l3 & all & l1 & l2 & l3 & all & l1 & l2 & l3 \\
\midrule
GPT-4.1 & 0.98 & 0.32 & 1.07 & 3.51 & 53.33 & 69.81 & 50.00 & 30.77 & 0.705 & 0.367 & 0.710 & 1.380 & 243K & 103K & 249K & 506K \\
Claude 3.7 Sonnet & 3.54 & 1.69 & 3.81 & 9.04 & 61.82 & 73.58 & 60.47 & 42.31 & 2.190 & 1.244 & 2.301 & 3.824 & 680K & 379K & 716K & 1,196K \\
Qwen3-235B-A22B & 0.22 & 0.12 & 0.30 & 0.27 & 27.27 & 37.74 & 22.09 & 23.08 & 0.040 & 0.082 & 0.091 & 0.093 & 72K & 53K & 81K & 76K \\
Qwen3-30B-A3B & 0.13 & 0.07 & 0.16 & $\infty$ & 17.58 & 30.19 & 15.12 & 0.00 & 0.023 & 0.022 & 0.024 & 0.022 & 65K & 61K & 70K & 60K \\
QwQ-32B & 0.23 & 0.15 & 0.26 & 0.49 & 22.42 & 30.19 & 20.93 & 11.54 & 0.120 & 0.102 & 0.126 & 0.135 & 142K & 129K & 148K & 149K \\
o1 & 3.66 & 1.96 & 3.62 & 12.66 & 52.12 & 67.92 & 50.00 & 26.92 & 1.908 & 1.328 & 1.812 & 3.408 & 69K & 47K & 66K & 127K \\
\bottomrule
\end{tabular}
}
\label{tab:model}
\end{table*}

Based on the results, we have several findings:
Claude 3.7 Sonnet achieves the highest accuracy on the GAIA benchmark (61.82\% overall) compared to GPT-4.1 (53.33\%), but its cost-of-pass is significantly higher (3.54 vs. 0.98). This indicates that current high-performing LLMs, when used as agent backbones, often sacrifice efficiency for better effectiveness, highlighting a critical trade-off in model design.
Sparse models like Qwen3-30B-A3B exhibit superior efficiency, with a low cost-of-pass (0.13 overall) despite modest accuracy (17.58\% overall). Given GAIA's challenging nature and Qwen3-30B-A3B's small activated parameter count (3B), such models may offer advantages for simpler agent tasks where efficiency is prioritized over raw performance. MoE-based sparse models, such as Qwen3-30B-A3B, leverage selective parameter activation to achieve remarkable efficiency, making them well-suited for resource-constrained agent tasks.
As task difficulty increases from Level 1 to Level 3, cost-of-pass rises dramatically across large reasoning models. For instance, Claude 3.7 Sonnet’s cost-of-pass increases from 1.69 to 9.04 (a 534\% surge) and OpenAI o1 from 1.96 to 12.66 (a 646\% surge), underscoring that efficiency deteriorates significantly on harder tasks especially for reasoning models, posing challenges for scaling LLMs to complex agent scenarios.
\begin{AIbox}{Reasoning Models' Efficiency Deterioration on Hard Task}
    As task difficulty escalates, cost-of-pass of reasoning models dramatically increases and efficiency significantly deteriorates, posing a formidable challenge for deploying these models in intricate agentic environments.

\end{AIbox}

\begin{figure}
    \centering
    \includegraphics[width=0.8\linewidth]{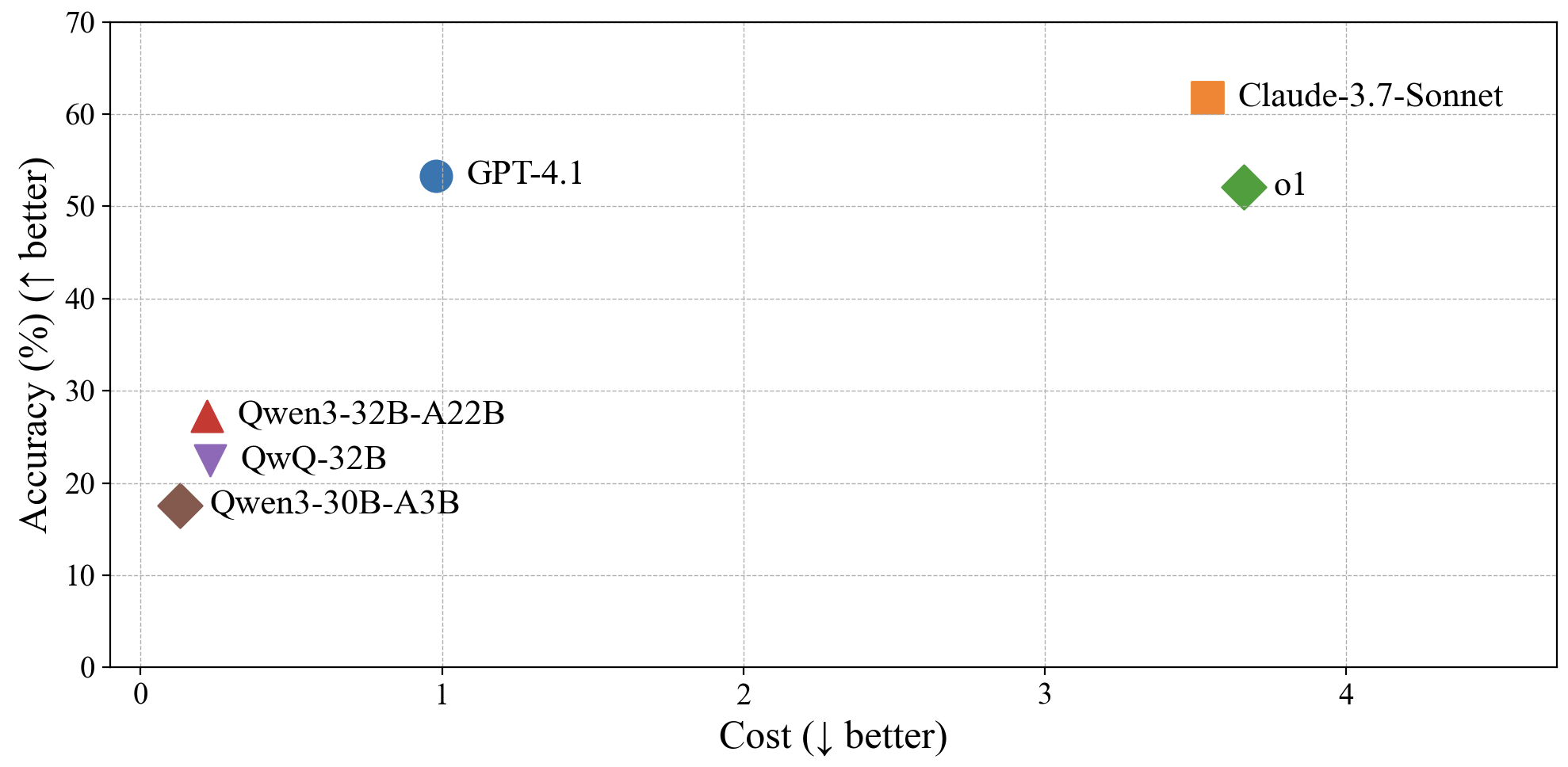}
    \caption{Performance of various backbone LLMs on the GAIA benchmark: Accuracy vs Cost.}
    \label{fig:enter-label}
\end{figure}

\subsection{Test-time Scaling Strategies}
Test-time scaling enhance models performance by leveraging multiple inference runs \citep{brown2024largelanguagemonkeysscaling,snell2024scalingllmtesttimecompute}, but these approaches typically require the model to be executed $N$ times, significantly increasing token consumption. We evaluate the common strategies, Best-of-$N$ (BoN) \citep{brown2024largelanguagemonkeysscaling}. At each step, $N$ possible actions are sampled and evaluated by a progress reward model (PRM). The action scored the highest is kept as the next action. We implement the PRM by promting GPT-4o. The prompt can be found in Appendix \ref{app:prompt}. We test $N \in {1, 2, 4}$,. The results are presented in Table \ref{tab:tts}.

\begin{table*}[h]
\centering
\caption{Best-of-N Performance under Different N in Test-time Scaling.}
\resizebox{\textwidth}{!}{
\begin{tabular}{c|cccc|cccc|cccc|cccc}
\toprule
\multirow{3}{*}{N} & \multicolumn{4}{c|}{Efficiency} & \multicolumn{4}{c|}{Effectiveness} & \multicolumn{8}{c}{Cost} \\
\cmidrule(lr){2-5} \cmidrule(lr){6-9} \cmidrule(lr){10-17}
& \multicolumn{4}{c|}{cost-of-pass$\downarrow$} & \multicolumn{4}{c|}{Acc.$\uparrow$} & \multicolumn{4}{c|}{Cost/\textdollar$\downarrow$} & \multicolumn{4}{c}{\#Tokens$\downarrow$} \\
\cmidrule(lr){2-5} \cmidrule(lr){6-9} \cmidrule(lr){10-13} \cmidrule(lr){14-17}
& all & l1 & l2 & l3 & all & l1 & l2 & l3 & all & l1 & l2 & l3 & all & l1 & l2 & l3 \\
\midrule
1 & 0.98 & 0.32 & 1.07 & 3.51 & 53.33 & 69.81 & 50.00 & 30.77 & 0.521 & 0.227 & 0.533 & 1.080 & 243K & 103K & 249K & 506K \\
2 & 1.17 & 0.54 & 1.18 & 3.70 & 54.55 & 62.26 & 56.98 & 30.77 & 0.639 & 0.336 & 0.675 & 1.138 & 298K & 153K & 317K & 533K \\
4 & 1.28 & 0.65 & 1.02 & 2.70 & 53.94 & 71.70 & 46.51 & 42.31 & 0.691 & 0.350 & 0.734 & 1.253 & 325K & 161K & 345K & 593K \\
\bottomrule
\end{tabular}
}
\label{tab:tts}
\end{table*}

We observe that increasing $N$ in Best-of-$N$ from 1 to 4 leads to a substantial rise in token consumption (from 243k to 325k). However, the performance improvement is marginal, with accuracy only slightly increasing from 53.33\% to 53.94\% when $N=4$ compared to $N=1$. Consequently, this results in a notable decrease in efficiency, with the cost-of-pass rising from 0.98 to 1.28.
\begin{AIbox}{Best-of-N Rises Cost But Marginal Gain}
    The marginal performance gains of BoN come at a disproportionate computational cost, highlighting the need for more efficient test-time scaling strategies in an agent setting.
\end{AIbox}

\subsection{Planning}
To enhance the agent's ability to handle long-horizon tasks, a planning module prior to execution is usually adopted. Planning can be regarded as a continuous task decomposition process \citep{huang2024understandingplanningllmagents}. To generally evaluate the impact of the planning module on efficiency, we adopt a simple and universal design.

Specifically, the agent is prompted to generate an explicit plan before taking any action. It then follows this plan step by step in a ReAct \citep{yao2023reactsynergizingreasoningacting} style. To allow for adaptability in dynamic environments, the plan is periodically revised: after every $N$ steps, the agent re-generates the plan based on the current context. The specific prompt used for planning is provided in Appendix \ref{app:prompt}.

In our experiments, we vary the maximum number of ReAct steps allowed totally, choosing from ${4, 8, 12}$. Additionally, we control the frequency of plan updates by setting the planning interval $N$ to ${1, 2, 4}$. The detailed results of these variations are summarized in Table \ref{tab:planning}.

\begin{table*}[h]
\centering
\caption{Results on different planning methods of agents. In the upper part, we keep the planning interval fixed at 1 and adjust the maximum steps. In the lower part, we keep the maximum steps fixed at 12 and adjust the planning interval.}
\resizebox{\textwidth}{!}{
\begin{tabular}{cc|cccc|cccc|cccc|cccc}
\toprule
\multirow{3}{*}{Max Steps} & \multirow{3}{*}{Plan Interval} & \multicolumn{4}{c|}{Efficiency} & \multicolumn{4}{c|}{Effectiveness} & \multicolumn{8}{c}{Cost} \\
\cmidrule(lr){3-6} \cmidrule(lr){7-10} \cmidrule(lr){11-18}
& & \multicolumn{4}{c|}{cost-of-pass$\downarrow$} & \multicolumn{4}{c|}{Acc.$\uparrow$} & \multicolumn{4}{c|}{Cost/\textdollar$\downarrow$} & \multicolumn{4}{c}{\#Tokens$\downarrow$} \\
\cmidrule(lr){3-6} \cmidrule(lr){7-10} \cmidrule(lr){11-14} \cmidrule(lr){15-18}
& & all & l1 & l2 & l3 & all & l1 & l2 & l3 & all & l1 & l2 & l3 & all & l1 & l2 & l3 \\
\midrule
12 & 1 & 0.98 & 0.32 & 1.07 & 3.51 & 53.33 & 69.81 & 50.00 & 30.77 & 0.705 & 0.367 & 0.710 & 1.380 & 243K & 103K & 249K & 506K \\
8 & 1 & 0.70 & 0.29 & 0.82 & 2.11 & 52.73 & 69.81 & 48.84 & 30.77 & 0.550 & 0.313 & 0.591 & 0.939 & 171K & 91K & 185K & 300K \\
4 & 1 & 0.48 & 0.29 & 0.51 & 2.05 & 41.82 & 58.49 & 39.53 & 15.38 & 0.199 & 0.167 & 0.200 & 0.316 & 88K & 73K & 89K & 141K \\
\midrule
12 & 1 & 0.98 & 0.32 & 1.07 & 3.51 & 53.33 & 69.81 & 50.00 & 30.77 & 0.705 & 0.367 & 0.710 & 1.380 & 243K & 103K & 249K & 506K \\
12 & 2 & 1.04 & 0.46 & 1.15 & 3.16 & 57.58 & 71.70 & 56.98 & 30.77 & 0.834 & 0.458 & 0.875 & 1.477 & 272K & 146K & 301K & 442K \\
12 & 4 & 1.01 & 0.40 & 0.97 & 3.91 & 53.33 & 66.04 & 52.33 & 30.77 & 0.772 & 0.389 & 0.725 & 1.710 & 249K & 118K & 235K & 563K \\
\bottomrule
\end{tabular}
}
\label{tab:planning}
\end{table*}

Increasing the maximum number of steps within a certain range significantly improves performance. For instance, when the maximum steps increase from 4 to 8, the accuracy rises from 58.49\% to 69.81\%. However, this also leads to a substantial increase in cost-of-pass (from 0.48 to 0.70). Beyond a certain threshold, further increasing the maximum steps does not enhance performance but continues to increase costs. 
\begin{AIbox}{Planning Complexity Control for Efficiency Optimization}
    Current models struggle with reasoning length regulation, often exhibiting overthinking that inflates costs when problems are insoluble. Moderate planning complexity significantly enhances efficiency.
\end{AIbox}

\subsection{Tool Using}
Incorporating external tools significantly enhances the agent's capabilities, especially in scenarios where neural networks alone fall short \citep{qin2024toollearningfoundationmodels}. However, this also introduces additional overhead. In this work, we focus primarily on the effectiveness and efficiency of using of a web browser \citep{nakano2022webgptbrowserassistedquestionansweringhuman} for two reasons: (1) it represents a widely adopted and general-purpose tool that enables agents to access real-time, up-to-date information across diverse domains, which is helpful for addressing a wide range of problems; (2) the use of browsing may have a significant impact on efficiency. Web pages often contain large amounts of text, multimedia, and interactive content, leading to high token consumption during content retrieval and processing. This is especially true considering that effective use of the browser may involve navigating multiple web pages. Therefore, we place particular emphasis on the efficiency and effectiveness of the browser use in our analysis.

In our experiments, we control for several factors related to browsing. These include: 
\begin{itemize}
    \item \textbf{Source of Web Content}: We evaluate the impact of sources by comparing a simple setting comprising only Google and Wikipedia against a complex setting that includes Google, Wikipedia, Bing, Baidu, and DuckDuckGo.
    \item \textbf{Web Page Processing Strategy}: We design three distinct approaches: (a) a crawler that retrieves only static elements, (b) a browser with basic processing, and (c) a browser with advanced operations, such as page-up and page-down interactions.
    \item \textbf{Number of Query Expansion}: the original user query is reformulated by the LLM to obtain a broader and more informative set of results, with expansion numbers to \{3, 5, 10\}.
\end{itemize}
 The results are presented in Table \ref{tab:tool}.

\begin{table*}[h]
\centering
\caption{Efficiency and effectiveness on different settings of tool using. We evaluate on different design on searching sources, browsing tools and the number of query rewritten.}
\resizebox{\textwidth}{!}{
\begin{tabular}{ccc|cccc|cccc|cccc|cccc}
\toprule
\multirow{3}{*}{Source} & \multirow{3}{*}{Tool} & \multirow{3}{*}{Search Num} & \multicolumn{4}{c|}{Efficiency} & \multicolumn{4}{c|}{Effectiveness} & \multicolumn{8}{c}{Cost} \\
\cmidrule(lr){4-7} \cmidrule(lr){8-11} \cmidrule(lr){12-19}
& & & \multicolumn{4}{c|}{cost-of-pass$\downarrow$} & \multicolumn{4}{c|}{Acc.$\uparrow$} & \multicolumn{4}{c|}{Cost/\textdollar$\downarrow$} & \multicolumn{4}{c}{\#Tokens$\downarrow$} \\
\cmidrule(lr){4-7} \cmidrule(lr){8-11} \cmidrule(lr){12-15} \cmidrule(lr){16-19}
&  &  & all & l1 & l2 & l3 & all & l1 & l2 & l3 & all & l1 & l2 & l3 & all & l1 & l2 & l3 \\
\midrule
Simple & Crawler & 10 & 1.32 & 0.53 & 1.42 & 4.49 & 53.33 & 69.81 & 50.00 & 30.77 & 0.705 & 0.367 & 0.710 & 1.380 & 243K & 103K & 249K & 506K \\
Multi & Crawler & 10 & 0.81 & 0.35 & 0.92 & 2.43 & 59.39 & 73.58 & 59.30 & 30.77 & 0.479 & 0.258 & 0.545 & 0.749 & 225K & 118K & 257K & 353K \\
\midrule
Simple & Browser-Complex & 10 & 0.88 & 0.42 & 0.85 & 4.18 & 49.09 & 62.26 & 50.00 & 19.23 & 0.431 & 0.259 & 0.424 & 0.804 & 199K & 118K & 196K & 374K \\
Simple & Browser-Simple & 10 & 1.59 & 0.64 & 1.75 & 5.60 & 58.18 & 73.58 & 56.98 & 30.77 & 0.927 & 0.468 & 0.996 & 1.724 & 327K & 155K & 350K &636K \\
\midrule
Simple & Crawler & 5 & 1.17 & 0.49 & 1.30 & 3.22 & 53.33 & 66.04 & 51.16 & 34.62 & 0.626 & 0.322 & 0.667 & 1.114 & 212K & 101K & 229K & 385K \\
Simple & Crawler & 3 & 1.31 & 0.61 & 1.33 & 3.97 & 49.09 & 60.38 & 47.67 & 30.77 & 0.641 & 0.366 & 0.635 & 1.220 & 215K & 114K & 219K & 405K \\
\bottomrule
\end{tabular}
}
\label{tab:tool}
\end{table*}

Increasing the number of search sources significantly enhances both effectiveness and efficiency. Specifically, the cost-of-pass decreases from 1.32 to 0.81, while accuracy improves from 53.33\% to 59.39\%.
Simpler browser operations, such as retrieving static elements or basic processing, outperform advanced operations (e.g., page-up and page-down interactions) in both effectiveness and efficiency, indicating that minimal processing strategies can achieve robust performance with lower computational overhead. Expanding the number of reformulated queries (from 3 to 10) consistently improves both effectiveness and efficiency, as broader query sets enable the retrieval of more relevant and informative results.
\begin{AIbox}{Tool Configurations Significantly Impact Efficiency and Effectiveness}
    Varying tool configurations, such as increasing search sources, simplifying browser operations, and expanding reformulated queries for web searching, demonstrably enhance both effectiveness and efficiency in information retrieval.
\end{AIbox}

\subsection{Memory}
Memory is a critical component for LLM-driven agent systems, enabling effective interaction with and learning from dynamic environments \citep{zhang2024surveymemorymechanismlarge,openai2025memory}. It supports key functionalities such as experience accumulation  and knowledge abstraction as the reasoning, while memory module also introduces extra cost. In our experiments, we design six memory configurations to evaluate their impact on effectiveness and efficiency to the whole system. Details of the prompts for the memory are provided in \ref{app:prompt}.
\begin{itemize}
    \item \textbf{Simple Memory}: Only historical observations and action are kept in the context window for a short context.
    \item \textbf{Summarized Memory}: At each step, all information—including observations, reasoning, and actions—is summarized by an LLM and embedded. These embeddings are stored in a vector database and retrieved based on cosine similarity, then concatenated into the prompt each step as an memory to replace the history of each step for a short context.
    \item \textbf{w/o Extra Memory}: Only the history of every step is kept in the context window while no extra memory are leveraged.
    \item \textbf{Extra Summarized Memory}: The memory is summarized exactly the same as Summarized Memory while concatenated into the prompt each step as an extra memory alongside the existing step history.
    \item \textbf{Extra Fixed Memory}: A piece of text with maxinum length is maintained and concatenated into the prompt at each step as long-term memory. It is initially generated by an LLM at the first step and subsequently updated by the LLM after every step.
    \item \textbf{Extra Hybrid Memory}: Concatenate both summarized and long memory approaches at each step to maintain more information.
\end{itemize}
We evaluated the effectiveness and efficiency by controlling different memory designs while keeping other settings the same, with results shown in Table \ref{tab:memory}.

\begin{table*}[h]
\centering
\caption{The impact on different design of Memory module on effectiveness and efficiency.}
\resizebox{\textwidth}{!}{
\begin{tabular}{c|cccc|cccc|cccc|cccc}
\toprule
\multirow{3}{*}{Memory} & \multicolumn{4}{c|}{Efficiency} & \multicolumn{4}{c|}{Effectiveness} & \multicolumn{8}{c}{Cost} \\
\cmidrule(lr){2-5} \cmidrule(lr){6-9} \cmidrule(lr){10-17}
& \multicolumn{4}{c|}{cost-of-pass$\downarrow$} & \multicolumn{4}{c|}{Acc.$\uparrow$} & \multicolumn{4}{c|}{Cost/\textdollar$\downarrow$} & \multicolumn{4}{c}{\#Tokens$\downarrow$} \\
\cmidrule(lr){2-5} \cmidrule(lr){6-9} \cmidrule(lr){10-13} \cmidrule(lr){14-17}
& all & l1 & l2 & l3 & all & l1 & l2 & l3 & all & l1 & l2 & l3 & all & l1 & l2 & l3 \\
\midrule
simple & 0.74 & 0.46 & 0.64 & 1.28 & 56.36 & 66.04 & 56.98 & 34.62 & 0.419 & 0.258 & 0.426 & 0.727 & 194K & 117K & 196K & 338K \\ 
summarized &1.52	&0.87	&1.43	&2.01	&51.52	&66.04	&48.84	&30.77	&0.782	&0.449	&0.942	&0.983	&367K	&226K	&437K	&441K \\ 
w/o extra & 0.98 & 0.32 & 1.07 & 3.51 & 53.33 & 69.81 & 50.00 & 30.77 & 0.521 & 0.227 & 0.533 & 1.080 & 243K & 103K & 249K & 506K \\
extra summarized & 1.08 & 0.60 & 1.05 & 2.99 & 52.73 & 60.38 & 53.49 & 34.62 & 0.567 & 0.364 & 0.561 & 1.036 & 236K & 146K & 234K & 442K \\
extra fixed & 1.04 & 0.52 & 1.00 & 3.53 & 53.94 & 64.15 & 54.65 & 30.77 & 0.561 & 0.331 & 0.544 & 1.088 & 240K & 135K & 234K & 472K \\
extra hybrid & 1.29 & 0.68 & 1.32 & 5.29 & 54.55 & 75.47 & 51.16 & 23.08 & 0.703 & 0.512 & 0.677 & 1.220 & 259K & 176K & 253K & 462K \\
\bottomrule
\end{tabular}
}
\label{tab:memory}
\end{table*}

Among the six memory configurations, Simple Memory, which retains only the agent's observations and actions, minimizes the context window size, resulting in the lowest computational cost. Surprisingly, this configuration also yields the best performance, improving from 53.33\% (No Extra Memory baseline) to 56.36\%, while reducing the cost-of-pass from 0.98 to 0.74.
In contrast, Summarized Memory, which also aims to shorten the context window, incurs the highest token consumption and computational cost. This may be due to its inability to accurately summarize past historical trajectories, requiring the model to make additional attempts to solve tasks.
(3) Additional memory designs, which augment the existing step history, provide marginal performance improvements but are outperformed by the Simple Memory configuration.
\begin{AIbox}{Simple Memory is Enough}
The Simple Memory design, retaining only the agent's observations and actions, is sufficient to achieve both effectiveness and efficiency.
\end{AIbox}

\subsection{Holistic Analysis of Component Impacts on Agent System}
In this section, we adopt a global perspective to examine how various components of the agent system influence its effectiveness and efficiency. Our analysis reveals that the choice of backbone exerts the most significant impact on the overall system performance. Additionally, the maximum number of steps an agent can execute and the usage of tools also play critical roles in determining performance. In contrast, the design of BoN and memory mechanisms has negligible effects on the model's effectiveness. However, redundant designs in these components may lead to increased computational costs.
\section{\method: Tricks of the Trade}
\begin{table}[h]
\centering
\caption{The Configuration of \method. The choice of each component is conducted by the observation from the previous empirical studies.}
\begin{tabular}{c|ccccccc}
\toprule
\textbf{Component} & \textbf{Backbone} & \textbf{Max Step} & \textbf{Plan Interval} & \textbf{Search Source} & \textbf{Search Num} & \textbf{BoN} & \textbf{Memory} \\
\midrule
\textbf{Settings} & GPT-4.1 & 8 & 1 & Multi & 5 & 1 & Simple \\
\bottomrule
\end{tabular}
\label{tab:efficient_agent}
\end{table}
In this section, we propose \method, an agent system comprising carefully selected components to achieve a great trade-off between effectiveness and efficiency. We demonstrate that by tuning the configuration based on empirical studies and selecting components that achieve a favorable trade-off between efficiency and effectiveness, the resulting agent system can maintain performance while significantly reducing costs. Specifically, for each component in the agent system, we adopt the configuration with the lowest cost-of-pass among those that do not lead to substantial performance degradation. The detailed configurations are provided in Table~\ref{tab:efficient_agent}.

\begin{table*}[h]
\centering
\caption{Results on Different Agents.}
\resizebox{\textwidth}{!}{
\begin{tabular}{c|cccc|cccc|cccc|cccc}
\toprule
\multirow{3}{*}{Agent} & \multicolumn{4}{c|}{Efficiency} & \multicolumn{4}{c|}{Effectiveness} & \multicolumn{8}{c}{Cost} \\
\cmidrule(lr){2-5} \cmidrule(lr){6-9} \cmidrule(lr){10-17}
& \multicolumn{4}{c|}{cost-of-pass$\downarrow$} & \multicolumn{4}{c|}{Acc.$\uparrow$} & \multicolumn{4}{c|}{Cost/\textdollar$\downarrow$} & \multicolumn{4}{c}{\#Tokens$\downarrow$} \\
\cmidrule(lr){2-5} \cmidrule(lr){6-9} \cmidrule(lr){10-13} \cmidrule(lr){14-17}
& all & l1 & l2 & l3 & all & l1 & l2 & l3 & all & l1 & l2 & l3 & all & l1 & l2 & l3 \\
\midrule
OWL & 0.75	&0.35	&0.80	&2.10	&53.33	&71.70	&50.00	&26.92	&0.398	&0.248	&0.402	&0.566	&189K	&119K	&204K	&281K \\
Smolagents & 5.82 & 3.21 & 6.46 & 13.37 & 53.33 & 62.26 & 54.65 & 30.77 & 3.104 & 2.000 & 3.528 & 4.115 & 146K & 88K & 172K & 183K \\
\textbf{Efficient Agent (ours)} & 0.55 & 0.37 & 0.54 & 1.71 & 51.52 & 62.26 & 52.33 & 26.92 & 0.285 & 0.228 & 0.280 & 0.461 & 127K & 101K & 125K & 206K \\
\bottomrule
\end{tabular}
}
\label{tab:framework}
\end{table*}

For comparison, we conduct a comparative analysis of some popular open-source agent systems, including OWL~\citep{owl2025} and SmolAgent~\citep{smolagents}. These systems offer diverse designs in terms of planning, memory, and tool-use integration, representing some of the most actively developed agent systems in the community.

The results are listed in Table~\ref{tab:framework}. By evaluating these frameworks under GAIA benchmark, we find that our \method achieves a cost reduction of 28.4\% while maintaining comparable performance.

\section{Related Work}
\subsection{LLM-driven Agents}
LLM-based agent technologies have demonstrated remarkable capabilities across a wide range of tasks, significantly spurring the rapid advancement of general agent systems. In recent years, increasing research efforts have been dedicated to building general agent systems capable of tackling complex reasoning, planning, and search tasks, with the aim of enhancing their adaptability and automation capabilities in real-world scenarios. By leveraging LLMs' strengths in context understanding, knowledge integration, and tool use, these systems have proven successful on multiple public benchmarks \citep{mialon2023gaiabenchmarkgeneralai,wei2025browsecompsimplechallengingbenchmark}, underscoring their immense potential for building general, versatile, and multi-agent collaborative systems.
\\
For instance, OpenAI's Deep Research agent achieved an average score of 67.36\% on the GAIA (General AI Assistants) benchmark \citep{mialon2023gaiabenchmarkgeneralai}, significantly outperforming traditional LLM-only approaches. Within the open-source community, the OWL (Optimized Workforce Learning), scored an impressive average of 69.7\% on the GAIA benchmark \citep{owl2025}, achieving state-of-the-art performance in the open-source domain. These compelling results on demanding benchmarks like GAIA clearly demonstrate the significant potential of LLM-based agent systems for handling intricate tasks that require sophisticated reasoning, planning, and effective tool utilization.
\subsection{Efficient NLP}
Since the advent of BERT \citep{devlin2019bertpretrainingdeepbidirectional}, the scale of language models has grown exponentially, leading to substantial increases in computational and energy costs during inference. To address this, a significant body of research has focused on enhancing NLP efficiency~\citep{xu-etal-2020-bert,zhou2020bert,xu-etal-2021-beyond,zhou-etal-2022-bert,zhou-etal-2023-modular,zhou2023efficientpromptingdynamicincontext,wang-etal-2023-efficientvlm,zhang2025uora}. For instance, DistilBERT \citep{sanh2020distilbertdistilledversionbert} leverages knowledge distillation to create a compact model from BERT, maintaining strong performance across NLP tasks with reduced size and faster inference. 

More recently, as large reasoning models advent, one line of inquiry explores methods to control model output length by estimating the likely token requirements for a given task, thereby promoting efficient generation. For example, Token-Budget-Aware LLM Reasoning\citep{han2024token} introduces the concept of a token budget. By incorporating a reasonable token budget into the prompt, this approach dynamically estimates the inference complexity for different problems, guiding the reasoning process to significantly reduce token consumption with only a marginal performance trade-off. 

In the domain of general intelligent agent systems, the collaboration of multiple LLM agents, while powerful for complex tasks, often introduces inefficiencies such as communication redundancy and resource wastage. To mitigate these, strategies like AgentPrune \citep{zhang2024cut} focus on optimizing communication by pruning superfluous messages from a spatio-temporal communication graph.
Complementing these efforts to improve multi-agent efficiency, other research, such as that leading to BudgetMLAgent \citep{gandhi2025budgetmlagentcosteffectivellmmultiagent} explores the use of tiered model architectures that strategically combine lower-cost and high-performance LLMs to achieve cost-effective systems.

While existing research has achieved significant results in communication topology optimization and overall cost control, a detailed analysis of the efficiency contribution and cost impact of individual modules within general agent systems remains relatively unexplored. Our work focuses on a deeper investigation into the factors influencing cost contributed by each specific module in general agent systems.

\section{Conclusion}
Building upon our investigation into the efficiency-effectiveness trade-off in LLM-driven agents, this paper makes the following key contributions. First, we provide a comprehensive analysis of the architectural choices and operational factors that contribute to the substantial economic overhead observed in contemporary agent systems. This analysis pinpoints specific areas where inefficiency commonly arises, laying the groundwork for more cost-conscious design. Second, guided by these insights, we introduce \method, a novel agent framework engineered for an optimal balance between task performance and computational cost. Through careful selection and integration of its components, \method dynamically adapts its complexity to the demands of the task at hand. Our extensive experiments on the challenging GAIA benchmark demonstrate the efficacy of our approach. Specifically, \method achieves 96.7\% of the state-of-the-art performance of OWL while drastically reducing the operational cost by xx times, resulting in a significant 28.4\% improvement in the cost-of-pass metric. This work underscores the critical importance of efficiency considerations in the design of next-generation agent systems and offers a practical pathway towards more scalable and economically viable real-world deployments. We believe our findings will spur further research into task-adaptive and resource-aware agent architectures, paving the way for more widespread adoption of these powerful AI technologies.
\newpage
\section*{Contributions}
\textbf{Core Contributors}
\begin{itemize}
    \item Ningning Wang
    \item Xavier Hu
\end{itemize}

\textbf{Contributors}
\begin{itemize}
    \item Pai Liu
    \item Yue Hou
    \item Heyuan Huang
    \item Shengyu Zhang
    \item Jian Yang
    \item Jiaheng Liu
    \item Ge Zhang
    \item Changwang Zhang
    \item Jun Wang
    \item Yuchen Eleanor Jiang
\end{itemize}

\textbf{Corresponding Authors}
\begin{itemize}
    \item He Zhu
    \item Wangchunshu Zhou
\end{itemize}

\clearpage

\bibliographystyle{unsrtnat}
\bibliography{efficient}

\begin{thebibliography}{52}
\providecommand{\natexlab}[1]{#1}
\providecommand{\url}[1]{\texttt{#1}}
\expandafter\ifx\csname urlstyle\endcsname\relax
  \providecommand{\doi}[1]{doi: #1}\else
  \providecommand{\doi}{doi: \begingroup \urlstyle{rm}\Url}\fi

\bibitem[Wu et~al.(2023)Wu, Bansal, Zhang, Wu, Li, Zhu, Jiang, Zhang, Zhang, Liu, Awadallah, White, Burger, and Wang]{wu2023autogen}
Qingyun Wu, Gagan Bansal, Jieyu Zhang, Yiran Wu, Beibin Li, Erkang Zhu, Li~Jiang, Xiaoyun Zhang, Shaokun Zhang, Jiale Liu, Ahmed~Hassan Awadallah, Ryen~W White, Doug Burger, and Chi Wang.
\newblock Autogen: Enabling next-gen llm applications via multi-agent conversation, 2023.
\newblock URL \url{https://arxiv.org/abs/2308.08155}.

\bibitem[Zhou et~al.(2023{\natexlab{a}})Zhou, Jiang, Li, Wu, Wang, Qiu, Zhang, Chen, Wu, Wang, Zhu, Chen, Zhang, Tang, Zhang, Chen, Cui, and Sachan]{zhou2023agents}
Wangchunshu Zhou, Yuchen~Eleanor Jiang, Long Li, Jialong Wu, Tiannan Wang, Shi Qiu, Jintian Zhang, Jing Chen, Ruipu Wu, Shuai Wang, Shiding Zhu, Jiyu Chen, Wentao Zhang, Xiangru Tang, Ningyu Zhang, Huajun Chen, Peng Cui, and Mrinmaya Sachan.
\newblock Agents: An open-source framework for autonomous language agents.
\newblock 2023{\natexlab{a}}.
\newblock URL \url{https://arxiv.org/abs/2309.07870}.

\bibitem[Zhou et~al.(2024)Zhou, Ou, Ding, Li, Wu, Wang, Chen, Wang, Xu, Zhang, Chen, and Jiang]{zhou2024agents2}
Wangchunshu Zhou, Yixin Ou, Shengwei Ding, Long Li, Jialong Wu, Tiannan Wang, Jiamin Chen, Shuai Wang, Xiaohua Xu, Ningyu Zhang, Huajun Chen, and Yuchen~Eleanor Jiang.
\newblock Symbolic learning enables self-evolving agents.
\newblock 2024.
\newblock URL \url{https://arxiv.org/abs/2406.18532}.

\bibitem[Zhou et~al.(2023{\natexlab{b}})Zhou, Jiang, Cui, Wang, Xiao, Hou, Cotterell, and Sachan]{zhou2023recurrentgpt}
Wangchunshu Zhou, Yuchen~Eleanor Jiang, Peng Cui, Tiannan Wang, Zhenxin Xiao, Yifan Hou, Ryan Cotterell, and Mrinmaya Sachan.
\newblock Recurrentgpt: Interactive generation of (arbitrarily) long text, 2023{\natexlab{b}}.
\newblock URL \url{https://arxiv.org/abs/2305.13304}.

\bibitem[{Significant Gravitas}(2025)]{autogpt2025}
{Significant Gravitas}.
\newblock Autogpt, 2025.
\newblock URL \url{https://github.com/Significant-Gravitas/AutoGPT}.
\newblock Accessed: 2025-05-13.

\bibitem[Zhu et~al.(2025{\natexlab{a}})Zhu, Qin, Zhu, Huang, Guan, Xia, Yao, Li, Wang, Liu, Peng, Gui, Li, Liu, Jiang, Wang, Zhang, Tang, Zhang, Yang, Liu, Gao, Liu, and Zhou]{zhu2025oagentsempiricalstudybuilding}
He~Zhu, Tianrui Qin, King Zhu, Heyuan Huang, Yeyi Guan, Jinxiang Xia, Yi~Yao, Hanhao Li, Ningning Wang, Pai Liu, Tianhao Peng, Xin Gui, Xiaowan Li, Yuhui Liu, Yuchen~Eleanor Jiang, Jun Wang, Changwang Zhang, Xiangru Tang, Ge~Zhang, Jian Yang, Minghao Liu, Xitong Gao, Jiaheng Liu, and Wangchunshu Zhou.
\newblock Oagents: An empirical study of building effective agents, 2025{\natexlab{a}}.
\newblock URL \url{https://arxiv.org/abs/2506.15741}.

\bibitem[Zhu et~al.(2025{\natexlab{b}})Zhu, Li, Wu, Xing, Ma, Tang, Liu, Yang, Liu, Jiang, Zhang, Lin, Wang, Zhang, and Zhou]{zhu2025scalingtesttimecomputellm}
King Zhu, Hanhao Li, Siwei Wu, Tianshun Xing, Dehua Ma, Xiangru Tang, Minghao Liu, Jian Yang, Jiaheng Liu, Yuchen~Eleanor Jiang, Changwang Zhang, Chenghua Lin, Jun Wang, Ge~Zhang, and Wangchunshu Zhou.
\newblock Scaling test-time compute for llm agents, 2025{\natexlab{b}}.
\newblock URL \url{https://arxiv.org/abs/2506.12928}.

\bibitem[Hu et~al.(2024{\natexlab{a}})Hu, Zhao, Wei, Chai, Ma, Wang, Wang, Su, Xu, Zhu, Cheng, Yuan, Li, Kuang, Yang, Yang, and Wu]{hu2024infiagentdabenchevaluatingagentsdata}
Xueyu Hu, Ziyu Zhao, Shuang Wei, Ziwei Chai, Qianli Ma, Guoyin Wang, Xuwu Wang, Jing Su, Jingjing Xu, Ming Zhu, Yao Cheng, Jianbo Yuan, Jiwei Li, Kun Kuang, Yang Yang, Hongxia Yang, and Fei Wu.
\newblock Infiagent-dabench: Evaluating agents on data analysis tasks, 2024{\natexlab{a}}.
\newblock URL \url{https://arxiv.org/abs/2401.05507}.

\bibitem[Hu et~al.(2024{\natexlab{b}})Hu, Xiong, Yi, Wei, Xiao, Chen, Ye, Tao, Zhou, Zhao, et~al.]{hu2024agents}
Xueyu Hu, Tao Xiong, Biao Yi, Zishu Wei, Ruixuan Xiao, Yurun Chen, Jiasheng Ye, Meiling Tao, Xiangxin Zhou, Ziyu Zhao, et~al.
\newblock Os agents: A survey on mllm-based agents for computer, phone and browser use, 2024{\natexlab{b}}.

\bibitem[Shi et~al.(2025)Shi, Cao, Chen, Sun, Li, Lu, Dong, Qin, Zhu, Liu, Yang, Zhang, Liu, Zhang, Wang, Jiang, and Zhou]{shi2025taskcraftautomatedgenerationagentic}
Dingfeng Shi, Jingyi Cao, Qianben Chen, Weichen Sun, Weizhen Li, Hongxuan Lu, Fangchen Dong, Tianrui Qin, King Zhu, Minghao Liu, Jian Yang, Ge~Zhang, Jiaheng Liu, Changwang Zhang, Jun Wang, Yuchen~Eleanor Jiang, and Wangchunshu Zhou.
\newblock Taskcraft: Automated generation of agentic tasks, 2025.
\newblock URL \url{https://arxiv.org/abs/2506.10055}.

\bibitem[Devlin et~al.(2019)Devlin, Chang, Lee, and Toutanova]{devlin2019bertpretrainingdeepbidirectional}
Jacob Devlin, Ming-Wei Chang, Kenton Lee, and Kristina Toutanova.
\newblock Bert: Pre-training of deep bidirectional transformers for language understanding, 2019.
\newblock URL \url{https://arxiv.org/abs/1810.04805}.

\bibitem[Achiam et~al.(2023)Achiam, Adler, Agarwal, Ahmad, Akkaya, Aleman, Almeida, Altenschmidt, Altman, Anadkat, et~al.]{achiam2023gpt}
Josh Achiam, Steven Adler, Sandhini Agarwal, Lama Ahmad, Ilge Akkaya, Florencia~Leoni Aleman, Diogo Almeida, Janko Altenschmidt, Sam Altman, Shyamal Anadkat, et~al.
\newblock Gpt-4 technical report.
\newblock \emph{arXiv preprint arXiv:2303.08774}, 2023.

\bibitem[Kaplan et~al.(2020)Kaplan, McCandlish, Henighan, Brown, Chess, Child, Gray, Radford, Wu, and Amodei]{kaplan2020scalinglawsneurallanguage}
Jared Kaplan, Sam McCandlish, Tom Henighan, Tom~B. Brown, Benjamin Chess, Rewon Child, Scott Gray, Alec Radford, Jeffrey Wu, and Dario Amodei.
\newblock Scaling laws for neural language models, 2020.
\newblock URL \url{https://arxiv.org/abs/2001.08361}.

\bibitem[Hoffmann et~al.(2022)Hoffmann, Borgeaud, Mensch, Buchatskaya, Cai, Rutherford, de~Las~Casas, Hendricks, Welbl, Clark, Hennigan, Noland, Millican, van~den Driessche, Damoc, Guy, Osindero, Simonyan, Elsen, Rae, Vinyals, and Sifre]{hoffmann2022trainingcomputeoptimallargelanguage}
Jordan Hoffmann, Sebastian Borgeaud, Arthur Mensch, Elena Buchatskaya, Trevor Cai, Eliza Rutherford, Diego de~Las~Casas, Lisa~Anne Hendricks, Johannes Welbl, Aidan Clark, Tom Hennigan, Eric Noland, Katie Millican, George van~den Driessche, Bogdan Damoc, Aurelia Guy, Simon Osindero, Karen Simonyan, Erich Elsen, Jack~W. Rae, Oriol Vinyals, and Laurent Sifre.
\newblock Training compute-optimal large language models, 2022.
\newblock URL \url{https://arxiv.org/abs/2203.15556}.

\bibitem[Sanh et~al.(2020)Sanh, Debut, Chaumond, and Wolf]{sanh2020distilbertdistilledversionbert}
Victor Sanh, Lysandre Debut, Julien Chaumond, and Thomas Wolf.
\newblock Distilbert, a distilled version of bert: smaller, faster, cheaper and lighter, 2020.
\newblock URL \url{https://arxiv.org/abs/1910.01108}.

\bibitem[Abdin et~al.(2024)Abdin, Aneja, Behl, Bubeck, Eldan, Gunasekar, Harrison, Hewett, Javaheripi, Kauffmann, Lee, Lee, Li, Liu, Mendes, Nguyen, Price, de~Rosa, Saarikivi, Salim, Shah, Wang, Ward, Wu, Yu, Zhang, and Zhang]{abdin2024phi4technicalreport}
Marah Abdin, Jyoti Aneja, Harkirat Behl, Sébastien Bubeck, Ronen Eldan, Suriya Gunasekar, Michael Harrison, Russell~J. Hewett, Mojan Javaheripi, Piero Kauffmann, James~R. Lee, Yin~Tat Lee, Yuanzhi Li, Weishung Liu, Caio C.~T. Mendes, Anh Nguyen, Eric Price, Gustavo de~Rosa, Olli Saarikivi, Adil Salim, Shital Shah, Xin Wang, Rachel Ward, Yue Wu, Dingli Yu, Cyril Zhang, and Yi~Zhang.
\newblock Phi-4 technical report, 2024.
\newblock URL \url{https://arxiv.org/abs/2412.08905}.

\bibitem[{OpenAI}(2025{\natexlab{a}})]{openai_deep_research}
{OpenAI}.
\newblock Deep research, 2025{\natexlab{a}}.
\newblock URL \url{https://openai.com/index/deep-research-system-card/}.
\newblock Accessed: 2025-05-13.

\bibitem[{Monica.ai}(2025)]{manus}
{Monica.ai}.
\newblock manus, 2025.
\newblock URL \url{https://manus.im/app}.
\newblock Accessed: 2025-05-13.

\bibitem[Mialon et~al.(2023)Mialon, Fourrier, Swift, Wolf, LeCun, and Scialom]{mialon2023gaiabenchmarkgeneralai}
Grégoire Mialon, Clémentine Fourrier, Craig Swift, Thomas Wolf, Yann LeCun, and Thomas Scialom.
\newblock Gaia: a benchmark for general ai assistants, 2023.
\newblock URL \url{https://arxiv.org/abs/2311.12983}.

\bibitem[Erol et~al.(2025{\natexlab{a}})Erol, El, Suzgun, Yuksekgonul, and Zou]{erol2025costofpasseconomicframeworkevaluating}
Mehmet~Hamza Erol, Batu El, Mirac Suzgun, Mert Yuksekgonul, and James Zou.
\newblock Cost-of-pass: An economic framework for evaluating language models, 2025{\natexlab{a}}.
\newblock URL \url{https://arxiv.org/abs/2504.13359}.

\bibitem[{CAMEL-AI.org}(2025)]{owl2025}
{CAMEL-AI.org}.
\newblock Owl: Optimized workforce learning for general multi-agent assistance in real-world task automation.
\newblock \url{https://github.com/camel-ai/owl}, 2025.
\newblock Accessed: 2025-03-07.

\bibitem[Wang et~al.(2024)Wang, Jain, Zhang, Ray, Kumar, and Athiwaratkun]{wang2024reasoning}
Junlin Wang, Siddhartha Jain, Dejiao Zhang, Baishakhi Ray, Varun Kumar, and Ben Athiwaratkun.
\newblock Reasoning in token economies: Budget-aware evaluation of llm reasoning strategies.
\newblock In \emph{Proceedings of the 2024 Conference on Empirical Methods in Natural Language Processing}, pages 19916--19939, 2024.

\bibitem[Huang et~al.(2024)Huang, Liu, Chen, Wang, Wang, Lian, Wang, Tang, and Chen]{huang2024understandingplanningllmagents}
Xu~Huang, Weiwen Liu, Xiaolong Chen, Xingmei Wang, Hao Wang, Defu Lian, Yasheng Wang, Ruiming Tang, and Enhong Chen.
\newblock Understanding the planning of llm agents: A survey, 2024.
\newblock URL \url{https://arxiv.org/abs/2402.02716}.

\bibitem[Qin et~al.(2024)Qin, Hu, Lin, Chen, Ding, Cui, Zeng, Huang, Xiao, Han, Fung, Su, Wang, Qian, Tian, Zhu, Liang, Shen, Xu, Zhang, Ye, Li, Tang, Yi, Zhu, Dai, Yan, Cong, Lu, Zhao, Huang, Yan, Han, Sun, Li, Phang, Yang, Wu, Ji, Liu, and Sun]{qin2024toollearningfoundationmodels}
Yujia Qin, Shengding Hu, Yankai Lin, Weize Chen, Ning Ding, Ganqu Cui, Zheni Zeng, Yufei Huang, Chaojun Xiao, Chi Han, Yi~Ren Fung, Yusheng Su, Huadong Wang, Cheng Qian, Runchu Tian, Kunlun Zhu, Shihao Liang, Xingyu Shen, Bokai Xu, Zhen Zhang, Yining Ye, Bowen Li, Ziwei Tang, Jing Yi, Yuzhang Zhu, Zhenning Dai, Lan Yan, Xin Cong, Yaxi Lu, Weilin Zhao, Yuxiang Huang, Junxi Yan, Xu~Han, Xian Sun, Dahai Li, Jason Phang, Cheng Yang, Tongshuang Wu, Heng Ji, Zhiyuan Liu, and Maosong Sun.
\newblock Tool learning with foundation models, 2024.
\newblock URL \url{https://arxiv.org/abs/2304.08354}.

\bibitem[Zhang et~al.(2024{\natexlab{a}})Zhang, Bo, Ma, Li, Chen, Dai, Zhu, Dong, and Wen]{zhang2024surveymemorymechanismlarge}
Zeyu Zhang, Xiaohe Bo, Chen Ma, Rui Li, Xu~Chen, Quanyu Dai, Jieming Zhu, Zhenhua Dong, and Ji-Rong Wen.
\newblock A survey on the memory mechanism of large language model based agents, 2024{\natexlab{a}}.
\newblock URL \url{https://arxiv.org/abs/2404.13501}.

\bibitem[Snell et~al.(2024)Snell, Lee, Xu, and Kumar]{snell2024scalingllmtesttimecompute}
Charlie Snell, Jaehoon Lee, Kelvin Xu, and Aviral Kumar.
\newblock Scaling llm test-time compute optimally can be more effective than scaling model parameters, 2024.
\newblock URL \url{https://arxiv.org/abs/2408.03314}.

\bibitem[Erol et~al.(2025{\natexlab{b}})Erol, El, Suzgun, Yuksekgonul, and Zou]{erol2025cost}
Mehmet~Hamza Erol, Batu El, Mirac Suzgun, Mert Yuksekgonul, and James Zou.
\newblock Cost-of-pass: An economic framework for evaluating language models.
\newblock \emph{arXiv preprint arXiv:2504.13359}, 2025{\natexlab{b}}.

\bibitem[Jaech et~al.(2024)Jaech, Kalai, Lerer, Richardson, El-Kishky, Low, Helyar, Madry, Beutel, Carney, et~al.]{jaech2024openai}
Aaron Jaech, Adam Kalai, Adam Lerer, Adam Richardson, Ahmed El-Kishky, Aiden Low, Alec Helyar, Aleksander Madry, Alex Beutel, Alex Carney, et~al.
\newblock Openai o1 system card.
\newblock \emph{arXiv preprint arXiv:2412.16720}, 2024.

\bibitem[Guo et~al.(2025)Guo, Yang, Zhang, Song, Zhang, Xu, Zhu, Ma, Wang, Bi, et~al.]{guo2025deepseek}
Daya Guo, Dejian Yang, Haowei Zhang, Junxiao Song, Ruoyu Zhang, Runxin Xu, Qihao Zhu, Shirong Ma, Peiyi Wang, Xiao Bi, et~al.
\newblock Deepseek-r1: Incentivizing reasoning capability in llms via reinforcement learning.
\newblock \emph{arXiv preprint arXiv:2501.12948}, 2025.

\bibitem[Wei et~al.(2022)Wei, Wang, Schuurmans, Bosma, Xia, Chi, Le, Zhou, et~al.]{wei2022chain}
Jason Wei, Xuezhi Wang, Dale Schuurmans, Maarten Bosma, Fei Xia, Ed~Chi, Quoc~V Le, Denny Zhou, et~al.
\newblock Chain-of-thought prompting elicits reasoning in large language models.
\newblock \emph{Advances in neural information processing systems}, 35:\penalty0 24824--24837, 2022.

\bibitem[Chen et~al.(2025)Chen, Xu, Liang, He, Pang, Yu, Song, Liu, Zhou, Zhang, Wang, Tu, Mi, and Yu]{chen2025think23overthinkingo1like}
Xingyu Chen, Jiahao Xu, Tian Liang, Zhiwei He, Jianhui Pang, Dian Yu, Linfeng Song, Qiuzhi Liu, Mengfei Zhou, Zhuosheng Zhang, Rui Wang, Zhaopeng Tu, Haitao Mi, and Dong Yu.
\newblock Do not think that much for 2+3=? on the overthinking of o1-like llms, 2025.
\newblock URL \url{https://arxiv.org/abs/2412.21187}.

\bibitem[{OpenAI}(2025{\natexlab{b}})]{gpt41}
{OpenAI}.
\newblock {GPT-4.1: An Advanced Multimodal AI Model}, 2025{\natexlab{b}}.
\newblock Developed by OpenAI, available at \url{https://openai.com}.

\bibitem[{Anthropic}(2025)]{Anthropic2025Claude37Sonnet}
{Anthropic}.
\newblock Claude 3.7 sonnet.
\newblock \url{https://www.anthropic.com/news/claude-3-7-sonnet}, February 2025.
\newblock Accessed: 2025-05-11.

\bibitem[Yang et~al.(2025)Yang, Li, Yang, Zhang, Hui, Zheng, Yu, Gao, Huang, Lv, et~al.]{yang2025qwen3}
An~Yang, Anfeng Li, Baosong Yang, Beichen Zhang, Binyuan Hui, Bo~Zheng, Bowen Yu, Chang Gao, Chengen Huang, Chenxu Lv, et~al.
\newblock Qwen3 technical report.
\newblock \emph{arXiv preprint arXiv:2505.09388}, 2025.

\bibitem[Team(2025)]{qwq32b}
Qwen Team.
\newblock Qwq-32b: Embracing the power of reinforcement learning, March 2025.
\newblock URL \url{https://qwenlm.github.io/blog/qwq-32b/}.

\bibitem[Brown et~al.(2024)Brown, Juravsky, Ehrlich, Clark, Le, Ré, and Mirhoseini]{brown2024largelanguagemonkeysscaling}
Bradley Brown, Jordan Juravsky, Ryan Ehrlich, Ronald Clark, Quoc~V. Le, Christopher Ré, and Azalia Mirhoseini.
\newblock Large language monkeys: Scaling inference compute with repeated sampling, 2024.
\newblock URL \url{https://arxiv.org/abs/2407.21787}.

\bibitem[Yao et~al.(2023)Yao, Zhao, Yu, Du, Shafran, Narasimhan, and Cao]{yao2023reactsynergizingreasoningacting}
Shunyu Yao, Jeffrey Zhao, Dian Yu, Nan Du, Izhak Shafran, Karthik Narasimhan, and Yuan Cao.
\newblock React: Synergizing reasoning and acting in language models, 2023.
\newblock URL \url{https://arxiv.org/abs/2210.03629}.

\bibitem[Nakano et~al.(2022)Nakano, Hilton, Balaji, Wu, Ouyang, Kim, Hesse, Jain, Kosaraju, Saunders, Jiang, Cobbe, Eloundou, Krueger, Button, Knight, Chess, and Schulman]{nakano2022webgptbrowserassistedquestionansweringhuman}
Reiichiro Nakano, Jacob Hilton, Suchir Balaji, Jeff Wu, Long Ouyang, Christina Kim, Christopher Hesse, Shantanu Jain, Vineet Kosaraju, William Saunders, Xu~Jiang, Karl Cobbe, Tyna Eloundou, Gretchen Krueger, Kevin Button, Matthew Knight, Benjamin Chess, and John Schulman.
\newblock Webgpt: Browser-assisted question-answering with human feedback, 2022.
\newblock URL \url{https://arxiv.org/abs/2112.09332}.

\bibitem[{OpenAI}(2025{\natexlab{c}})]{openai2025memory}
{OpenAI}.
\newblock Memory and new controls for {ChatGPT}, February 2025{\natexlab{c}}.
\newblock URL \url{https://openai.com/index/memory-and-new-controls-for-chatgpt/}.
\newblock Accessed: 2025-05-14.

\bibitem[Roucher et~al.(2025)Roucher, del Moral, Wolf, von Werra, and Kaunismäki]{smolagents}
Aymeric Roucher, Albert~Villanova del Moral, Thomas Wolf, Leandro von Werra, and Erik Kaunismäki.
\newblock `smolagents`: a smol library to build great agentic systems.
\newblock \url{https://github.com/huggingface/smolagents}, 2025.

\bibitem[Wei et~al.(2025)Wei, Sun, Papay, McKinney, Han, Fulford, Chung, Passos, Fedus, and Glaese]{wei2025browsecompsimplechallengingbenchmark}
Jason Wei, Zhiqing Sun, Spencer Papay, Scott McKinney, Jeffrey Han, Isa Fulford, Hyung~Won Chung, Alex~Tachard Passos, William Fedus, and Amelia Glaese.
\newblock Browsecomp: A simple yet challenging benchmark for browsing agents, 2025.
\newblock URL \url{https://arxiv.org/abs/2504.12516}.

\bibitem[Xu et~al.(2020)Xu, Zhou, Ge, Wei, and Zhou]{xu-etal-2020-bert}
Canwen Xu, Wangchunshu Zhou, Tao Ge, Furu Wei, and Ming Zhou.
\newblock {BERT}-of-theseus: Compressing {BERT} by progressive module replacing.
\newblock In \emph{Proceedings of the 2020 Conference on Empirical Methods in Natural Language Processing (EMNLP)}, pages 7859--7869, Online, November 2020. Association for Computational Linguistics.
\newblock URL \url{https://www.aclweb.org/anthology/2020.emnlp-main.633}.

\bibitem[Zhou et~al.(2020)Zhou, Xu, Ge, McAuley, Xu, and Wei]{zhou2020bert}
Wangchunshu Zhou, Canwen Xu, Tao Ge, Julian McAuley, Ke~Xu, and Furu Wei.
\newblock Bert loses patience: Fast and robust inference with early exit.
\newblock In \emph{Advances in Neural Information Processing Systems}, volume~33, pages 18330--18341. Curran Associates, Inc., 2020.
\newblock URL \url{https://proceedings.neurips.cc/paper/2020/file/d4dd111a4fd973394238aca5c05bebe3-Paper.pdf}.

\bibitem[Xu et~al.(2021)Xu, Zhou, Ge, Xu, McAuley, and Wei]{xu-etal-2021-beyond}
Canwen Xu, Wangchunshu Zhou, Tao Ge, Ke~Xu, Julian McAuley, and Furu Wei.
\newblock Beyond preserved accuracy: Evaluating loyalty and robustness of {BERT} compression.
\newblock In Marie-Francine Moens, Xuanjing Huang, Lucia Specia, and Scott Wen-tau Yih, editors, \emph{Proceedings of the 2021 Conference on Empirical Methods in Natural Language Processing}, pages 10653--10659, Online and Punta Cana, Dominican Republic, November 2021. Association for Computational Linguistics.
\newblock \doi{10.18653/v1/2021.emnlp-main.832}.
\newblock URL \url{https://aclanthology.org/2021.emnlp-main.832/}.

\bibitem[Zhou et~al.(2022)Zhou, Xu, and McAuley]{zhou-etal-2022-bert}
Wangchunshu Zhou, Canwen Xu, and Julian McAuley.
\newblock {BERT} learns to teach: Knowledge distillation with meta learning.
\newblock In Smaranda Muresan, Preslav Nakov, and Aline Villavicencio, editors, \emph{Proceedings of the 60th Annual Meeting of the Association for Computational Linguistics (Volume 1: Long Papers)}, pages 7037--7049, Dublin, Ireland, May 2022. Association for Computational Linguistics.
\newblock \doi{10.18653/v1/2022.acl-long.485}.
\newblock URL \url{https://aclanthology.org/2022.acl-long.485/}.

\bibitem[Zhou et~al.(2023{\natexlab{c}})Zhou, Le~Bras, and Choi]{zhou-etal-2023-modular}
Wangchunshu Zhou, Ronan Le~Bras, and Yejin Choi.
\newblock Modular transformers: Compressing transformers into modularized layers for flexible efficient inference.
\newblock In Anna Rogers, Jordan Boyd-Graber, and Naoaki Okazaki, editors, \emph{Findings of the Association for Computational Linguistics: ACL 2023}, pages 10452--10465, Toronto, Canada, July 2023{\natexlab{c}}. Association for Computational Linguistics.
\newblock \doi{10.18653/v1/2023.findings-acl.664}.
\newblock URL \url{https://aclanthology.org/2023.findings-acl.664/}.

\bibitem[Zhou et~al.(2023{\natexlab{d}})Zhou, Jiang, Cotterell, and Sachan]{zhou2023efficientpromptingdynamicincontext}
Wangchunshu Zhou, Yuchen~Eleanor Jiang, Ryan Cotterell, and Mrinmaya Sachan.
\newblock Efficient prompting via dynamic in-context learning, 2023{\natexlab{d}}.
\newblock URL \url{https://arxiv.org/abs/2305.11170}.

\bibitem[Wang et~al.(2023)Wang, Zhou, Zeng, and Zhang]{wang-etal-2023-efficientvlm}
Tiannan Wang, Wangchunshu Zhou, Yan Zeng, and Xinsong Zhang.
\newblock {E}fficient{VLM}: Fast and accurate vision-language models via knowledge distillation and modal-adaptive pruning.
\newblock In Anna Rogers, Jordan Boyd-Graber, and Naoaki Okazaki, editors, \emph{Findings of the Association for Computational Linguistics: ACL 2023}, pages 13899--13913, Toronto, Canada, July 2023. Association for Computational Linguistics.
\newblock \doi{10.18653/v1/2023.findings-acl.873}.
\newblock URL \url{https://aclanthology.org/2023.findings-acl.873/}.

\bibitem[Zhang et~al.(2025)Zhang, Zhao, Yang, Zhong, Guan, Cao, and Wang]{zhang2025uora}
Xueyan Zhang, Jinman Zhao, Zhifei Yang, Yibo Zhong, Shuhao Guan, Linbo Cao, and Yining Wang.
\newblock Uora: Uniform orthogonal reinitialization adaptation in parameter-efficient fine-tuning of large models.
\newblock 2025.
\newblock URL \url{https://arxiv.org/abs/2505.20154}.

\bibitem[Han et~al.(2024)Han, Wang, Fang, Zhao, Ma, and Chen]{han2024token}
Tingxu Han, Zhenting Wang, Chunrong Fang, Shiyu Zhao, Shiqing Ma, and Zhenyu Chen.
\newblock Token-budget-aware llm reasoning.
\newblock \emph{arXiv preprint arXiv:2412.18547}, 2024.

\bibitem[Zhang et~al.(2024{\natexlab{b}})Zhang, Yue, Li, Yun, Wan, Wang, Cheng, Yu, and Chen]{zhang2024cut}
Guibin Zhang, Yanwei Yue, Zhixun Li, Sukwon Yun, Guancheng Wan, Kun Wang, Dawei Cheng, Jeffrey~Xu Yu, and Tianlong Chen.
\newblock Cut the crap: An economical communication pipeline for llm-based multi-agent systems.
\newblock \emph{arXiv preprint arXiv:2410.02506}, 2024{\natexlab{b}}.

\bibitem[Gandhi et~al.(2025)Gandhi, Patwardhan, Vig, and Shroff]{gandhi2025budgetmlagentcosteffectivellmmultiagent}
Shubham Gandhi, Manasi Patwardhan, Lovekesh Vig, and Gautam Shroff.
\newblock Budgetmlagent: A cost-effective llm multi-agent system for automating machine learning tasks, 2025.
\newblock URL \url{https://arxiv.org/abs/2411.07464}.

\end{thebibliography}

\clearpage

\beginappendix

\section{Default Setup}
Table~\ref{tab:default_setup} details the default setup for conducting the experiment.

\begin{table}[h]
\centering
\caption{The default setup for the experiment.}
\begin{tabular}{c|ccccccc}
\toprule
\textbf{Component} & \textbf{Backbone} & \textbf{Max Step} & \textbf{Plan Interval} & \textbf{Search Source} & \textbf{Search Num} & \textbf{BoN} & \textbf{Memory} \\
\midrule
\textbf{Settings} & GPT-4.1 & 12 & 1 & Simple & 10 & 1 & Simple \\
\bottomrule
\end{tabular}
\label{tab:default_setup}
\end{table}

\section{Prompt}
\label{app:prompt}

\subsection{Memory}
\resizebox{0.98\textwidth}{!}{
\begin{promptbox}[Memory Prompt]{ogreen}
You are an expert in agent memory management, specializing in leveraging the Memory Summarization, the Memory Retrieval, and the Long-term Memory to boost agent reasoning.

\textbf{Memory Summarization:}
\begin{itemize}
    \item Summarize the following text which is the execution content of the agent at the current step: \{memory of current step\}.
    \item Highlight the key points to assist the agent in better reasoning during subsequent steps.
    \item Additionally, you must provide optimization suggestions for the next step.
\end{itemize}

\textbf{Long-term Memory:}
\begin{itemize}
    \item Here is the agent's execution content from the previous step: \{memory of previous step\}. 
    \item Here is the long-term memory formed by summarizing the agent's historical execution content: \{long term memory\}. 
    \item Please combine the agent's previous execution content and the existing long-term memory, summarize them while highlighting the key points, and form a new long-term memory to help the agent reason better in subsequent steps.
\end{itemize}

\textbf{Input:}
\begin{itemize}
    \item Agent's execution content at current step: \{memory of current step\}.
    
    \item Agent's execution content at previous step: \{memory of previous step\}.
    
    \item Agent's historical execution content: \{long term memory\}.
\end{itemize}

\textbf{Output:}
\begin{itemize}
    \item \textbf{Memory Summarization:} A point-by-point summary of agent's current execution step and optimization suggestions.

    \item \textbf{Memory Retrieval:} The retrieval of the most relevant historical steps.
    
    \item \textbf{Long-term Memory:} An ongoing updated memory for recording the agent's long-term historical steps.
\end{itemize}

\end{promptbox}
}

\subsection{Test-Time Scaling}

\resizebox{0.98\textwidth}{!}{
\begin{promptbox}[PRM-score Evaluation Prompt]{ogreen}
\textbf{Evaluation Guidelines:}
\begin{itemize}
    \item \textbf{Objective:}
    \begin{itemize}
        \item You will evaluate a candidate ActionStep node, which includes the following fields:
        \begin{itemize}
            \item \texttt{step\_number}: Depth of this step within the TTS search tree.
            \item \texttt{observations}: Observations recorded after executing this action.
            \item \texttt{action\_output}: Direct output resulting from this action.
            \item \texttt{model\_output}: Raw LLM output that led to this action.
            \item \texttt{error}: Any encountered errors (can be None).
            \item \texttt{score}: Previously assigned score (for reference only).
            \item \texttt{previous\_steps}: The history of earlier steps, including TaskStep and PlanningStep, along with the trajectory of ActionSteps leading to the current state.
        \end{itemize}
        \item Your goal is to judge how promising this ActionStep is for advancing toward the user's task, using your independent judgment while considering the continuity and logical flow of the ActionStep sequence, including the historical context.
    \end{itemize}

    \item \textbf{Evaluation Criteria:}
    \begin{itemize}
        \item \textbf{Progress Toward Goal:}
        \begin{itemize}
            \item Assess whether the \texttt{action\_output} clearly and tangibly advances the overall task.
            \item Reward meaningful progress or valuable new information.
            \item Penalize irrelevant actions or weak impact.
        \end{itemize}
        \item \textbf{Error and Stability:}
        \begin{itemize}
            \item Penalize based on the severity of errors:
            \begin{itemize}
                \item Fatal/blocking errors: 0-1 points.
                \item Significant errors: 1-3 points.
                \item Minor or recoverable errors: 3-5 points.
            \end{itemize}
            \item Reduce the score if the \texttt{model\_output} is ambiguous or unstable.
        \end{itemize}
        \item \textbf{TTS Efficiency:}
        \begin{itemize}
            \item Reward actions that contribute efficiently toward reaching the goal.
            \item Penalize redundant or repetitive actions without meaningful progress.
        \end{itemize}
        \item \textbf{Reflection Usage:}
        \begin{itemize}
            \item Reward active utilization of \texttt{reflection} to improve upon past mistakes.
            \item Penalize ignoring reflection insights.
        \end{itemize}
        \item \textbf{Loop Detection:}
        \begin{itemize}
            \item Detect loops or repetitions compared to previous steps.
            \item Identify true loops and penalize based on severity.
        \end{itemize}
        \item \textbf{Contextual Awareness:}
        \begin{itemize}
            \item Infer alignment with previous \texttt{PlanningStep} and \texttt{TaskStep}.
            \item Ensure consistency with the TTS strategy and penalize deviations.
        \end{itemize}
    \end{itemize}

    \item \textbf{Scoring Criteria:}
    \begin{itemize}
        \item \texttt{9-10}: Clearly advances the goal; highly efficient; strong reflection use; no loops.
        \item \texttt{7-8}: Good advancement; minor inefficiencies; clear reflection use; minimal loop risk.
        \item \texttt{5-6}: Moderate progress; limited efficiency; moderate reflection use; mild repetition risks.
        \item \texttt{3-4}: Poor advancement; inefficient; weak reflection use; noticeable loop risks.
        \item \texttt{1-2}: Minimal advancement; repetitive actions; true loops; significant errors.
        \item \texttt{0}: Severe issues: explicit loops, critical errors, or complete irrelevance to the task context.
    \end{itemize}

    \item \textbf{Final Evaluation Output:}
    You must provide your evaluation in valid JSON format with the following structure:
    \begin{quote}
    \texttt{
    \{
  "analysis": "Detailed analysis addressing     progress, TTS efficiency, reflection usage, loop detection, contextual alignment with PlanningStep/TaskStep, error severity, and overall action quality.",
  "score": [integer between 0-10]
    \}
    }
    \end{quote}
    \end{itemize}
\end{promptbox}
}

\resizebox{0.98\textwidth}{!}{
\begin{promptbox}[PRM-list Evaluation Prompt]{ogreen}
\scriptsize 
\textbf{Evaluation Guidelines:}
\begin{itemize}
    \item \textbf{Objective:}
    \begin{itemize}
        \item You will evaluate N candidate trajectories, each representing a series of nodes in a search tree. Each trajectory contains the following:
        \begin{itemize}
            \item \texttt{step\_number}: Depth of the node in the trajectory.
            \item \texttt{observations}: Observations recorded at each step of the trajectory.
            \item \texttt{action\_output}: Direct action output at each step.
            \item \texttt{model\_output}: Raw model output (LLM).
            \item \texttt{error}: Any errors encountered (can be None).
            \item \texttt{score}: Previously calculated score (if available).
            \item \texttt{previous\_steps}: The history of earlier steps, including TaskStep and PlanningStep, with the trajectory of ActionSteps leading to the current state.
        \end{itemize}
        \item Your goal is to evaluate each trajectory holistically, considering how well it progresses toward solving the user's task. Select the trajectory that most effectively achieves this goal.
    \end{itemize}

    \item \textbf{Evaluation Criteria:}
    \begin{itemize}
        \item \textbf{Progress Toward Goal:}
        \begin{itemize}
            \item Assess how well each trajectory advances the task at hand, considering both the individual node's progress and the overall progression of the entire trajectory.
            \item Reward trajectories that demonstrate tangible and meaningful progress toward the goal.
            \item Penalize trajectories with weak actions or minimal/no advancement.
        \end{itemize}
        \item \textbf{Trajectory Efficiency:}
        \begin{itemize}
            \item Evaluate how efficiently each trajectory progresses toward the goal, considering the depth and complexity of the steps.
            \item Favor trajectories that achieve significant progress with fewer steps.
            \item Consider the overall value-to-depth ratio when comparing trajectories of different lengths.
            \item Reward efficient exploration of the search space.
        \end{itemize}
        \item \textbf{Loop Detection:}
        \begin{itemize}
            \item Detect loops or repetitions within each trajectory, especially those related to previous steps.
            \item \textbf{Loop types:}
            \begin{itemize}
                \item \texttt{Real Loops}: Identical nodes (observations, action output, and model output) that do not add value to the trajectory.
                \item \texttt{Benign Repetitions}: Similar strategies with variations yielding additional progress.
            \end{itemize}
            \item Heavily penalize trajectories with real loops.
            \item Slight penalties for benign repetitions if they lead to meaningful improvements.
        \end{itemize}
        \item \textbf{Error and Stability:}
        \begin{itemize}
            \item Evaluate the severity of errors encountered in each trajectory and penalize based on their impact on progression.
            \item \textbf{Error Severity:}
            \begin{itemize}
                \item Fatal/Blocking Errors: Major penalty.
                \item Significant Errors: Moderate penalty.
                \item Minor/Recoverable Issues: Minor penalty.
            \end{itemize}
            \item Penalize unstable or unclear model outputs.
            \item Consider how errors affect the overall trajectory's ability to move toward the goal.
        \end{itemize}
        \item \textbf{Overall Trajectory Quality:}
        \begin{itemize}
            \item Evaluate the coherence and overall quality of the trajectory.
            \item Consider the logical sequence of steps and the exploration-exploitation balance.
            \item Evaluate the final node's closeness to achieving the goal.
            \item Reward trajectories that make consistent progress and demonstrate coherent planning.
        \end{itemize}
    \end{itemize}

    \item \textbf{Final Output Format:}
    Provide your evaluation in the following JSON format. Select the best trajectory and provide a detailed analysis explaining why it is the most promising trajectory.
    \begin{quote}
    \texttt{
    \{
  "index": [integer],  \# Index of the best trajectory
  "analysis": "Detailed analysis addressing progress, efficiency, reflection usage, loop detection, error severity, and overall trajectory quality."
    \}
    }
    \end{quote}
    \end{itemize}

\end{promptbox}
}

\end{document}